%% file: main.tex
\documentclass[10pt,twocolumn,letterpaper]{article}

\usepackage{cvpr}      

\input{preamble}

\definecolor{cvprblue}{rgb}{0.21,0.49,0.74}
\usepackage[pagebackref,breaklinks,colorlinks,citecolor=cvprblue]{hyperref}

\def\paperID{9363} 
\def\confName{CVPR}

\usepackage{algorithm}
\usepackage{algorithmic}

\title{VectorTalker: SVG Talking Face Generation with Progressive Vectorisation}

\author{Hao Hu\textsuperscript{1} \and Xuan Wang\textsuperscript{2} \and Jingxiang Sun\textsuperscript{3} \and Yanbo Fan\textsuperscript{2} \and Yu Guo\textsuperscript{1} \and Caigui Jiang\textsuperscript{1}}

\begin{document}

\twocolumn[{
\renewcommand\twocolumn[1][]{#1}
\maketitle

\begin{center}
    \vspace{-15pt}
    \includegraphics[width=1.0\linewidth]{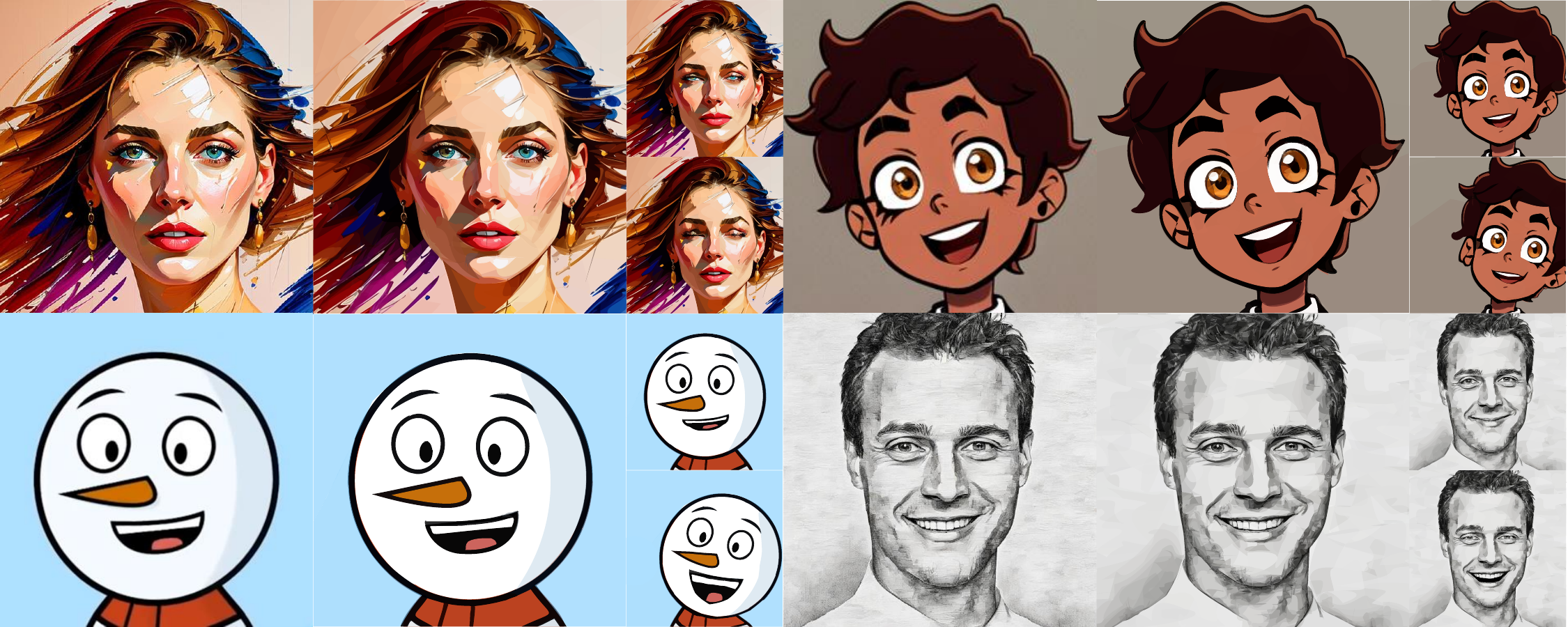}
    \vspace{-17pt}
    \captionsetup{type=figure}
    \caption{Image vectorization and SVG animation results. We can faithfully reconstruct SVG for different styles of raster portraits and provide vivid animation effects.
    }
    \label{fig:teaser}
    \vspace{16pt}
\end{center}
}]

\maketitle
\input{sec/0_abstract}    
\input{sec/1_intro}
\input{sec/2_relatedwork}
\input{sec/3_method}
\input{sec/4_exp}
\input{sec/5_conclusion}

{
    \small
    \bibliographystyle{ieeenat_fullname}
    \bibliography{main}
}

\input{sec/X_suppl}


\end{document}

%% file: preamble.tex
\usepackage[dvipsnames]{xcolor}

\usepackage{graphicx}
\usepackage[inkscapelatex=false]{svg}


%% file: sec/0_abstract.tex
\begin{abstract}

%
High-fidelity and efficient audio-driven talking head generation has been a key research topic in computer graphics and computer vision.
In this work, we study vector image based audio-driven talking head generation.
Compared with directly animating the raster image that most widely used in existing works, vector image enjoys its excellent scalability being used for many applications.
There are two main challenges for vector image based talking head generation: the high-quality vector image reconstruction {\it w.r.t.} the source portrait image and the vivid animation {\it w.r.t.} the audio signal.
To address these, we propose a novel scalable vector graphic reconstruction and animation method, dubbed VectorTalker.
Specifically, for the high-fidelity reconstruction, VectorTalker hierarchically reconstructs the vector image in a coarse-to-fine manner.
For the vivid audio-driven facial animation, we propose to use facial landmarks as intermediate motion representation and propose an efficient landmark-driven vector image deformation module.
Our approach can handle various styles of portrait images within a unified framework, including Japanese manga, cartoon, and photorealistic images. 
We conduct extensive quantitative and qualitative evaluations and the experimental results demonstrate the superiority of VectorTalker in both vector graphic reconstruction and audio-driven animation.
\end{abstract}

%% file: sec/1_intro.tex
\section{Introduction}
\label{sec:intro}

We study one-shot audio-driven talking head generation~\cite{tf1,tf2,tf3,tf4,tf5,tf6,tf7,tf8,tf9,tf10,tf11}, which aims to animate a single portrait image with speech audio.
In recent years, the applications of audio-driven facial animation has become ubiquitous in various fields, such as digital human creation, video conferences, and game industry, etc.
For these applications, the high visual quality of the animated faces and the ability of scalability are very important.
State-of-the-art realizations conduct facial animation in the input raster image space, for example, by learning a warping field on the raster image according to the audio signal.
However, because of the complex structure of the facial area, learning a reasonable warping field for each pixel or latent feature from speech remains challenging and the animated results suffer from distortions and blurs.
Besides, the visual quality of the facial animation is restricted by the training resolutions.
For instance, the scaled-up operation of the animated raster images will lead to blur or distortion artifacts.

In this work, we study the high-fidelity facial animation in the context of vector graphcis.
Unlike raster images that are composed of individual pixels, vector images are composed of mathematical primitives such as lines, curves, and geometric shapes in a resolution-independent fashion.
That is, the vector images can be scaled up or down without loss of visual quality.
Such scalability property makes vector images ideal for various output media and resolution sizes, which is highly appealing to talking head applications.
Besides, vector images possess the characteristic of editability.
The primitives of vector images are defined by mathematical equations that are easy to edit, modify, and adjust, without compromising image quality.
This makes it possible for high-quality facial animations.
Inspired by these promising properties, we study vector graphics based talking head animation.
Given a source raster image and an audio clip, we propose to first reconstruct the vector image {\it w.r.t.} the source image, and then perform facial animation in the vector image space. 
To the best of our knowledge, we are the first to explore audio-driven facial animation in the vector image space.
There are two main challenges:
1) the high-quality vector image reconstruction and 2) the vivid facial animation according to the input audio information. 

We propose a novel scalable vector graphic reconstruction and animation method, dubbed {\it VectorTalker}, that addresses the above challenges. 
Firstly, given a raster image, we construct the vector image using path primitives composed of $L$ segments of cubic Bezier curves.
The shape and color of each primitive are optimized based on the reconstruction error {\it w.r.t.} the raster image.
For a high-fidelity reconstruction, we propose a novel progressive vectorization algorithm.
Specifically, we perform $l_0$ regularized image smoothing on the input raster image with $N$-1 levels of smooth strengths which yields smoothed images as targets and then perform the differentiable vectorization progressively. 
In each level we add paths initialized as circles under guidance of the reconstruction error between the optimized paths and the smoothed image from last level. To facilitate subsequent SVG animation, we implement a semantic hierarchical design to reconstruct vector images of the background layer, foreground layer and local layer respectively.
Secondly, to enable vivid audio-driven animation, we use facial landmarks as intermediate motion representation and perform Delaunay triangulation on the vector image based on facial landmarks.
We then predict the landmark deformation according to the audio information and deform each path primitives of the vector image accordingly.
As the Bezier curve may crosses multiple triangulations, a simple unified transformation may leads to distortion results,
We propose to split the curves at the intersection points of the triangulation for more accurate animation. 
We conduct extensive experiments by considering various categories of facial portraits and several state-of-the-art methods. 
Experimental results demonstrate the effectiveness of VectorTalker and the necessity of the proposed coarse-to-fine vectorization and vivid animation designs.

The contributions of this paper can be summarized as follows:
\begin{itemize}
    \item We propose the first method to consider vector graphics based one-shot audio-driven talking head generation.
    \item We propose a novel progressive vectorization algorithm that obtains superior vector image reconstruction.
    \item We propose an efficient audio-driven module for vivid vector-image animation.
    \item We conduct extensive experiments and demonstrate the excellent performance of the proposed method.
\end{itemize}

{
}

\begin{figure*}[t]
\centering
\includegraphics[width=\textwidth]{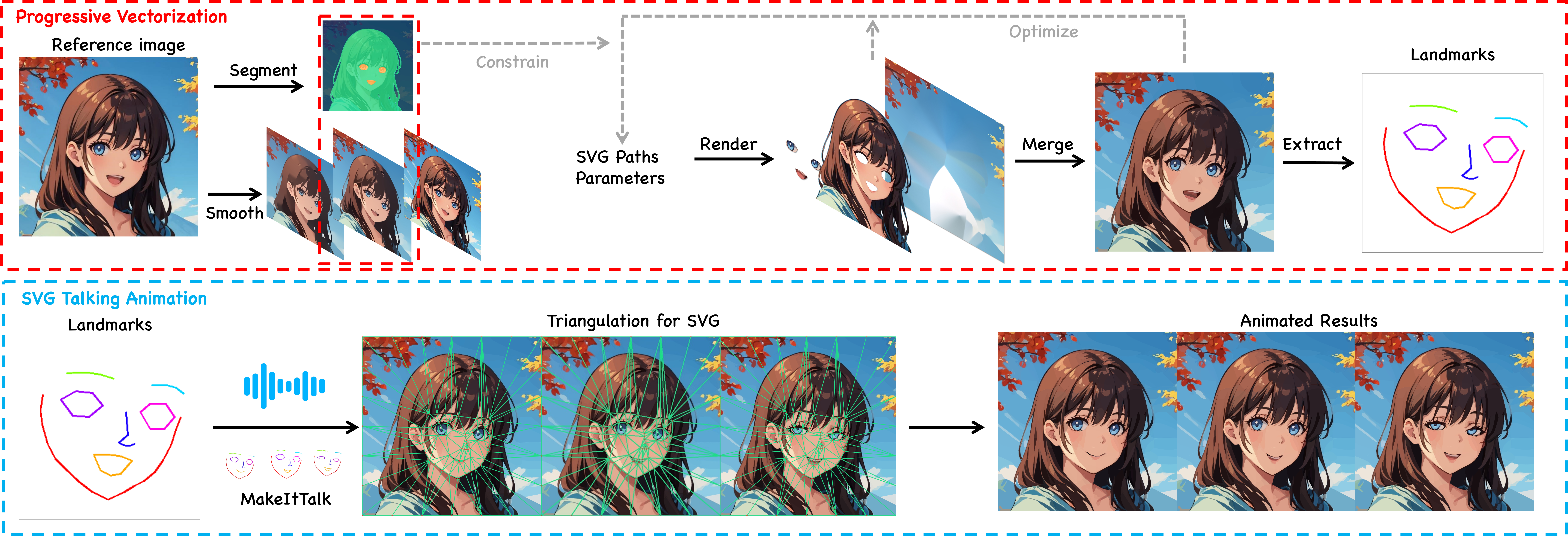}
\caption{Pipeline of our method. Given an input raster portrait image, our method first segment the image to get semantic mask (back, front, and local) and smooth the image to obtain different levels of smoothed representations as target image. Then we perform the differentiable vectorization to reconstruct three SVG layers constrained by semantic mask progressively and merge them to get final SVG result. For SVG talking animation, we extract landmarks and predict new ones from an audio clip to warp SVG paths by affine transformation.}
\label{fig:pipeline_v3}
\end{figure*}

\begin{figure}[t]
  \centering
   \includegraphics[width=0.5\textwidth]{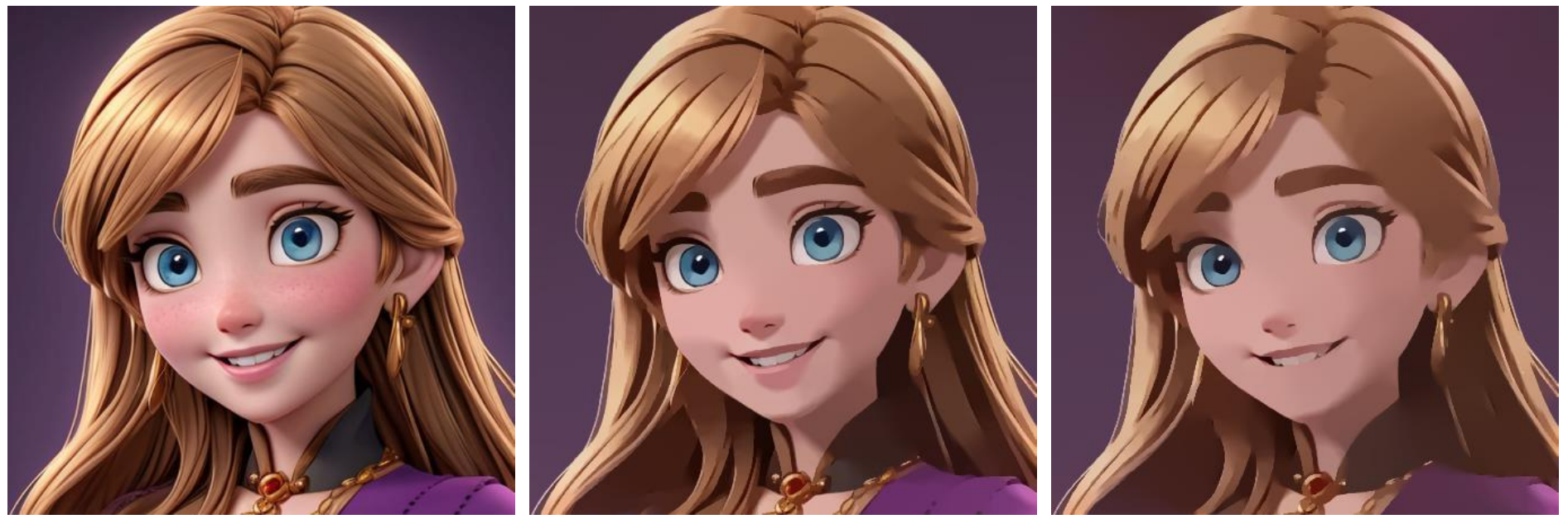}
   \caption{Image smoothing results. From left to right are the original image, light smoothed and heavy smoothed.}
   \label{fig:smooth}
\end{figure}

%% file: sec/2_relatedwork.tex
\section{Related Work} \label{sec:relatedwork}

\textbf{Image Vectorization.} Unlike raster images composed of pixels, vector graphics are composed of mathematical descriptions of geometric shapes. Most traditional vectorization methods~\cite{Potrace,Ardeco,Patch-Based} begin by segmenting the images into patches and then fit Bezier curves at the boundaries. To achieve smoother gradient colors, diffusion curves~\cite{DiffCurve1,DiffCurve2} and gradient meshes~\cite{GradMesh1,GradMesh2} are used to describe the colors. In recent years, The boom of deep learning  has promoted research on differentiable rendering of vector graphics so that raster-based algorithms can be used for vector generation. DiffVG~\cite{DiffVG} applies anti-aliasing to smooth vector graphics scene discontinuities and makes it differentiable. Im2Vec~\cite{Im2Vec} trains an encoder-decoder architecture from a raster dataset to predict a set of ordered closed vector paths. LIVE~\cite{LIVE} initializes component-wise path and optimizes the vector graph in a layer-wise manner and attemps to maintain the topological relationship of SVG, but it is only available for simple images and hard to handle complex textures.~\cite{tf16} work uses parameterization to describe facial attributes and transfer raster image to vector avatar by learning the mapping between both modalities, but it can only generate a single-style result.

\vspace{\baselineskip}

\noindent\textbf{Audio-driven Portrait Talking.} Most previous~\cite{tf1,tf2,tf3,tf4,tf5,tf6,tf7,tf8,tf9,tf10,tf11,tf12,tf13,tf14,tf15,SadTalker,MakeItTalk} works mainly focus on raster portrait talking.~\cite{tf1} train a RNN to learn mapping from audio features to mouth movements based on pre-built 3D facial models. AD-NeRF~\cite{tf11} trains two neural radiation fields by features extracted from audio to render more detailed results. SadTalker~\cite{SadTalker} proposed ExpNet and PoseVAE to predict 3D motion coefficients from audio, which are used as intermediate representations to generate videos. MakeItTalk~\cite{MakeItTalk} uses the facial landmark as the intermediate representation and maps pixels across frames via triangulation to make non-photorealistic portrait talk. To the best of our knowledge, there has been no prior work attempting to animate SVG portrait to talk. We are the first to apply talking generation methods used for raster to vector images.

%% file: sec/3_method.tex
\begin{figure*}[t]
  \centering
   \includegraphics[width=\linewidth]{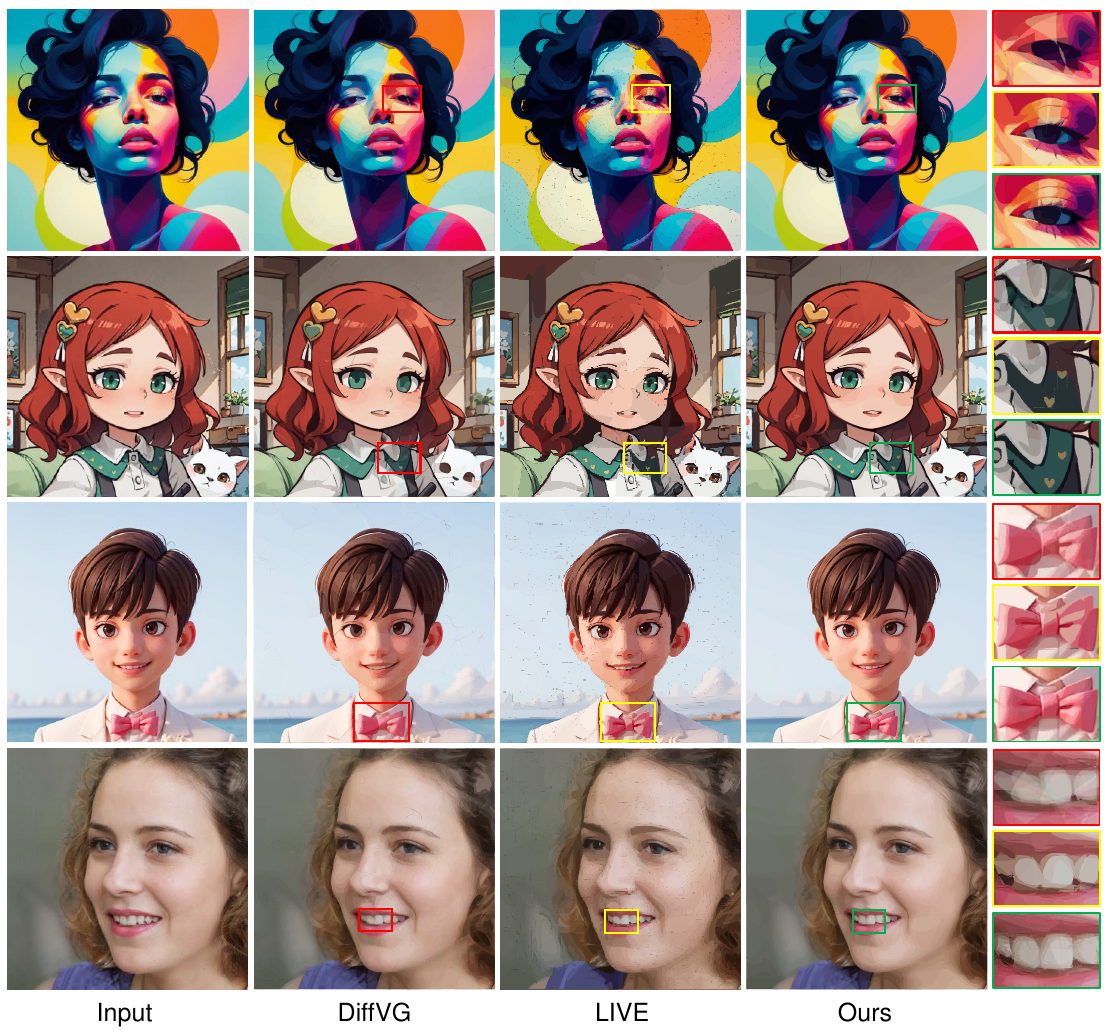}
   \caption{Qualitative results on SVG reconstruction. We compare our progressive vectorization algorithm with DiffVG and LIVE. The experiments illustrate that our method produces better results. We highlight the differences in the boxes on the right, and readers can zoom in for a clearer view.}
   \label{fig:compare1}
\end{figure*}

\begin{table}[htbp]
  \centering
  \begin{tabular}{ccccc}
    \toprule
    Method & MSE$\downarrow$ & LPIPS$\downarrow$ & PSNR$\uparrow$ & SSIM$\uparrow$ \\
    \midrule
    DiffVG & 0.00213 & 0.290 & 27.799 & 0.857 \\
    LIVE & 0.00194 & 0.269 & 28.291 & 0.864 \\
    Ours & \textbf{0.00131} & \textbf{0.258} & \textbf{29.835} & \textbf{0.886} \\
    \bottomrule
  \end{tabular}
  \caption{Quantitative evaluation using MSE, LPIPS, PNSR and SSIM on benchmark with different styles of portraits. Our method outperforms in all metrics.}
  \label{tab:rec}
\end{table}

\begin{figure}[t]
\centering
\includegraphics[width=0.5\textwidth]{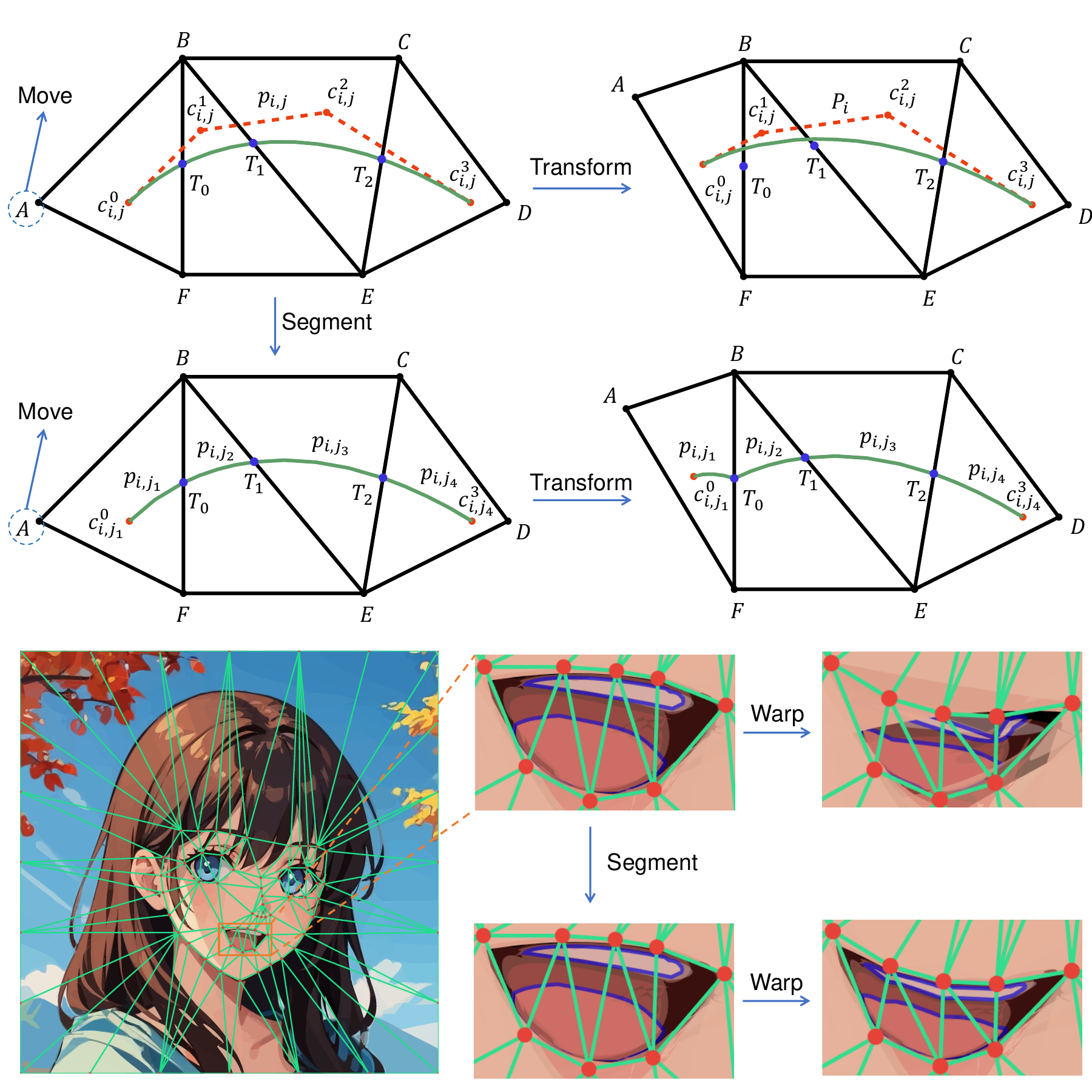}
\caption{Animation on SVG. The illustration of SVG animation and curve segmentation are shown. The abstraction illustration is shown above, and the qualitative display is shown below. If the paths are warped without curve segmentation, the result is unreasonable(lower) because a curve may span across multiple triangles and will be simultaneously affected by four affine transformations(upper). curve segmentation can maintain the shape of the curves and make paths be warped correctly. The abstracted illustration shows that the Bézier curve $\mathbf{p}_{i,j}$ and its four control points $\mathbf{c}^{*}_{i,j}, * = 0,1,2,3$ are scattered in 4 triangles. The affine transformations corresponding to triangles ${\Delta}AFB$, ${\Delta}BFE$, ${\Delta}BEC$ and ${\Delta}CED$ are $M_0$, $M_1$, $M_2$ and $M_3$. The entire deformed curve $\mathbf{p}_{i,j}$ will be simultaneously affected by four affine transformations, which is unfavorable for the SVG talking animation. Although we only move point A, the shape of the entire curve changes even if some parts are not within ${\Delta}AFB$. This issue can be effectively solve with the curve segmentaion by splitting $\mathbf{p}_{i,j}$ into $\mathbf{p}_{i,j_1}$, $\mathbf{p}_{i,j_2}$, $\mathbf{p}_{i,j_3}$ and $\mathbf{p}_{i,j_4}$.}
\label{fig:segcurve}
\end{figure}

\section{Method}
We aim to solve the problem of one-shot talking portrait generation in vector graphics which consist of multiple parametric paths as shown in~\cref{fig:pipeline_v3}. In this section, we expatriate the details of our proposed method, VectorTalker. The contents are organized as follows. First, the pipeline of reconstructing the SVG image from input raster images is described in ~\cref{sec:vectorization}. Then we present how to animate the reconstructed SVG portraits in~\cref{sec:animation}.

\textbf{Notations.} 
Let's start by defining the notations and concepts used in vector graphics and our approach. An SVG image is made up of graphic primitives or paths. In this method, we use a specific path $\mathbf{P}_i \in \mathcal{P}$ containing a series of third-order Bezier curves $\mathbf{p}_{i,j}$ which are closed. Each curve consists of four control points $\mathbf{c}^{*}_{i,j}$. Then we denote the SVG image as a path set $\mathcal{P}$, the raster image as $\mathbf{I}_*$, and the differentiable rasterization as the function $rast(\cdot)$.
In optimization, all the paths are initialized as circles with a defined radius $r$. We represent the stack of smoothing levels as $\mathcal{S}=\{\mathbf{S}^l|l=1, ..., N\}$. $L_f, L_m$ and $L_b$ are used to refer to the foreground, middle and background layers obtained by the foreground mask $\mathbf{M}_{fore}$ and the local mask $\mathbf{M}_{local}$ in the semantic layering. To create the animation, we extract a few facial key points, denoted by $\mathbf{k}_m$, from the input portrait. We then perform triangulation on these points to produce multiple triangles, represented by $F_t$. Using the offsets of the corresponding key points, we derive an affine transformation $M_t$ that is used to animate the paths in the image plane.

\subsection{Progressive Vectorization}\label{sec:vectorization} 

\noindent\textbf{Coarse-to-fine reconstruction.}
The first step of VectorTalker is to faithfully reconstruct the SVG image given an input raster portrait. SVG can be regarded as a parametric abstraction of the raster image. Therefore, a good vectorization of the raster image should capture most image structures having as few as possible paths. For a set of the optimizable paths, $\mathcal{P}$, random initialization in differentiable rendering usually leads to early convergence to tiny image structures and harms fidelity. 

Our approach is to use the progressive vectorization algorithm, which employs the coarse-to-fine strategy. The first step is to create a stack of smoothed images, called $\{\mathbf{S^l}\}$, using $N$-level smoothing. We do this by applying the $l_0$ regularized image smoothing to the input raster image as shown in~\cref{fig:smooth}. The smoothing strength is gradually increased for each level of the stack. For more information about the image smoothing algorithm, please see~\cite{l0}. By using $l_0$ regularized image smoothing, we can produce piece-wise constant images that depict the abstraction of the raster image in different degrees. This allows us to capture details in the different levels of the image. 

Given the $\mathcal{S}$, we perform the differentiable vectorization $rast(\cdot)$ progressively. With the decreasing of the smoothing level, we recursively initialize additional paths with a larger radius whose sampling positions are guided by the current error map and add them into the optimization variable stack. Then we optimize the set of paths $\mathcal{P}$ to fit the raster image in the current smoothing level by the MSE loss: $MSE(\mathbf{I}_s^{l} - rast(\mathcal{P}))$. Finally, the process above is repeated progressively across the levels of smoothing.~\cref{fig:compare1} show that our progressive vectorization significantly outperforms the baseline methods in image fidelity.


\noindent \textbf{Semantic Layering.}
To facilitate subsequent SVG animation, we implement a semantic hierarchical design. Specifically, we leverage the off-the-shelf segmentation model, e.g. SAM~\cite{SAM}, to extract the foreground mask $\mathbf{M}_{fore}$ and the local mask $\mathbf{M}_{local}$. The set of paths is separated into foreground, middle, and background layers using masks. Then, the sets of paths are merged semantically from back to front.

\begin{figure*}[h]
  \centering
   \includegraphics[width=\linewidth]{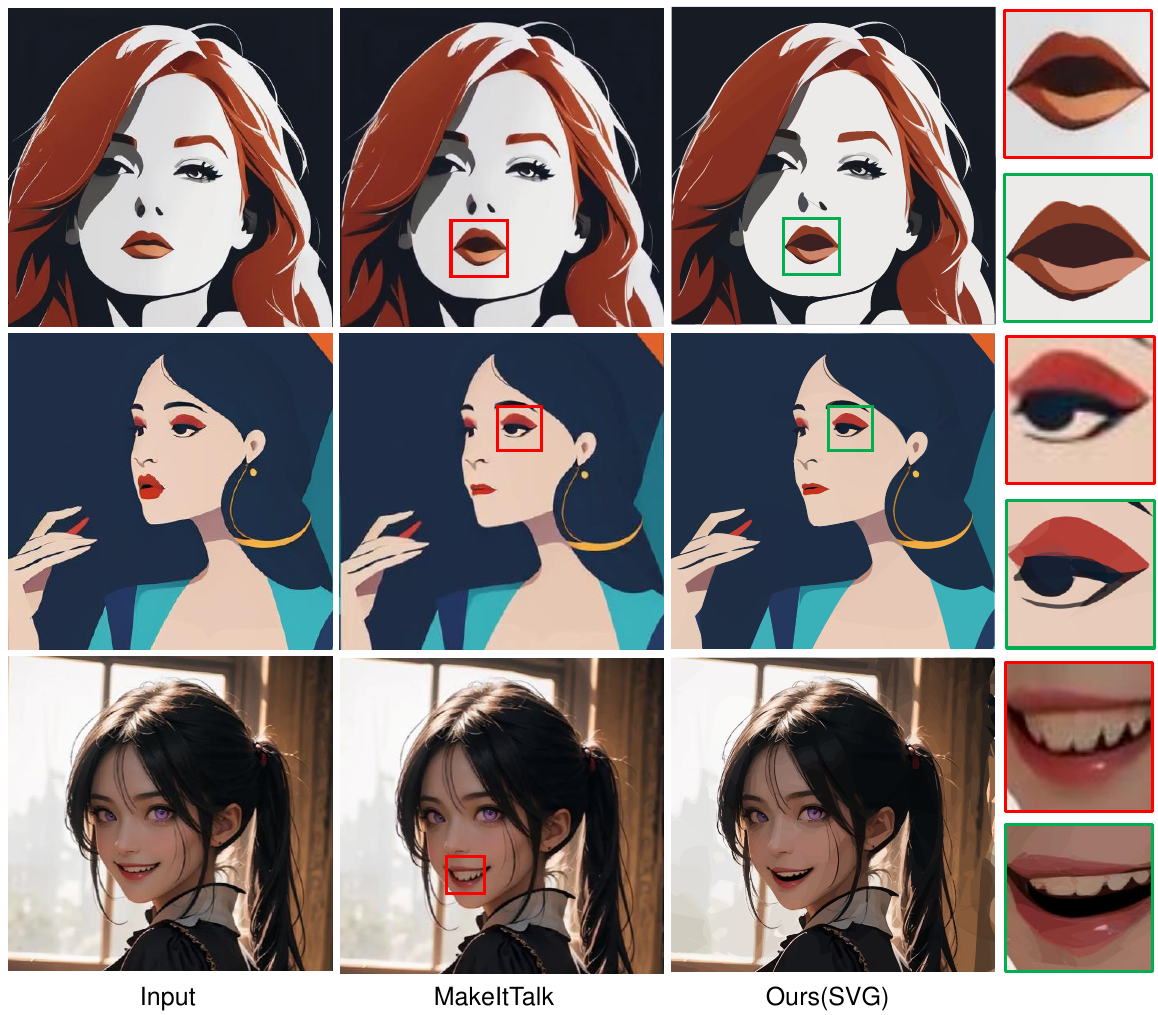}
   \caption{Qualitative comparison with MakeItTalk~\cite{}. The experiments illustrate that our results remain clear and sharp. Benefiting from the implementation of semantic layering in vectorization, the teeth perform reasonable motions.}
   \label{fig:ComMIT}
\end{figure*}

\begin{figure*}[h]
  \centering
   \includegraphics[width=\linewidth]{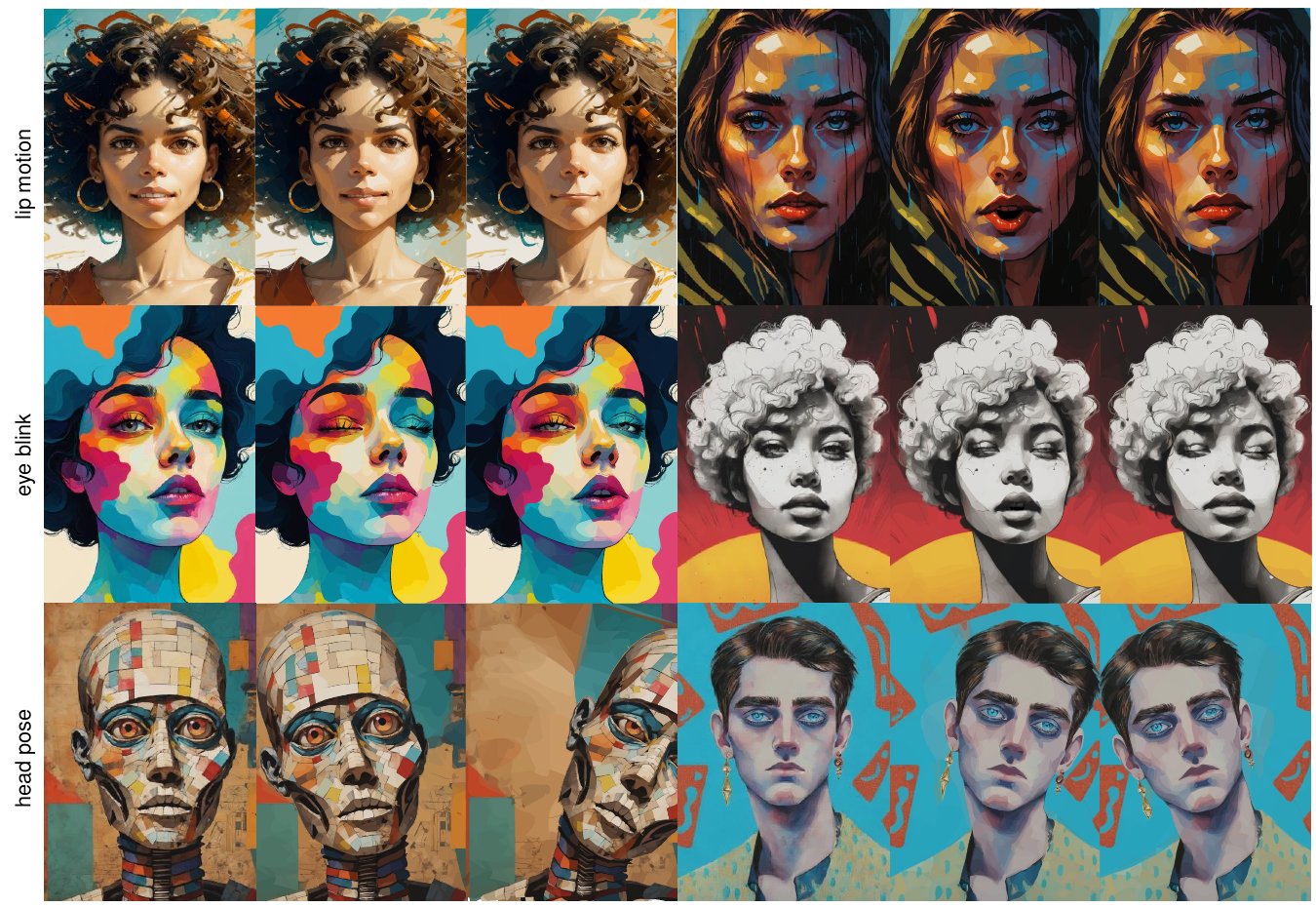}
   \caption{Our results of SVG animation. Our method makes SVG talk vividly. It allows the control of lip motion, eye blink and head poses.}
   \label{fig:BlinkTilt}
\end{figure*}

\subsection{Animating SVG Portraits}\label{sec:animation}
The aim of our work is to enable natural talking portraits by adjusting the position of the face landmarks. Similar to MakeItTalk~\cite{MakeItTalk}, but, instead of the raster image, we have to perform the animations on SVG images. After extracting the facial key points $\mathbf{k}_m$, Delaunay triangulation is then performed to divide the image into triangle patches based on the original landmarks. Although the coordinates of the triangle vertices change, the corresponding landmarks subscripts of the vertices remain fixed. We use facial landmarks as an intermediate representation between audio and visual animation, which transforms the subsequent animation process into the triangle transformation process. For each triangle, an affine transformation can be calculated based on the offsets and the original vertex coordinates. As long as the landmark topology remains unchanged, the texture on each triangle transfers across frames. In this paper, we adapt this approach, commonly used for raster images, into the case of vector graphics.

For the animation process of vector graphics, we use an off-the-shelf detector~\cite{foa} to predict the facial landmarks in the original image and perform Delaunay triangulation, following MakeItTalk~\cite{MakeItTalk}, and then we predict audio-driven sequence of landmarks, each of them determines the facial expression changes in a new frame. However, unlike raster images, SVG is composed of many paths of different shapes and colors and path is composed of several Bézier curves. Changing the shape of the paths requires changing the curves but simply performing affine transformation on the curve control points within each triangle may result in the path shape suffers unwanted distortion. The reason is that a Bézier curve may span across multiple triangles so that the control points are located within different triangles which often causes unreasonable motions. To address this issue, our solution is to segment all Bézier curves to ensure that each complete curve is inside only one triangle to avoid impact from other vertices. Specifically, we first calculate the intersections of each Bézier curve with all the line segments of the triangulation. 

Determining intersections only requires solving a cubic equation. By leveraging the properties of cubic Bézier curves, we can easily split the curve at the intersection points, ensuring that the shape of the path remains unchanged before and after the segmentation. When all control points of a curve are located within the same triangle, the entire curve undergoes the only affine transformation. As shown in the upper picture of~\cref{fig:segcurve}, if $\mathbf{p}_{i,j}$ is divided into 4 Bézier curves at intersections $T_0$, $T_1$ and $T_2$ and then we change the position of point A, only parts within ${\Delta}AFB$ will be changed and other parts will remain fixed. Although deformation after segmentation may destroy the smoothness of the curve, it is indispensable for subsequent animation. The lower picture of~\cref{fig:segcurve} shows the effect of curve segmentation on the SVG mouth region. If we directly warp paths without curves segmentation, the result is strange and messy. In contrast, segmenting a curve can maintain the shape of the path and make paths be warped correctly.

After completing the above steps, we can start driving SVG animation. However, as landmarks coordinates change, if we simply apply the same operation to all paths teeth and eyes may suffer unreasonable deformations and the background will also be distorted, such as the teeth becoming very large, the eyes being squeezed and the background Keep shaking. Thanks to the semantic layering we implement during reconstruction, our method can handle these problems well. Specifically, We can only warp the path in the foreground and local layers and keep the paths in the background layer fixed. For the eyes and mouth, we reselect landmarks for triangulation, so that the eyes move with the eye sockets and the mouth moves with the lower jaw. When blinking, the eyes can be naturally covered by the upper foreground layer without being squeezed and deformed and the teeth will not stretch to completely cover the entire mouth when lip opens. 

%% file: sec/4_exp.tex
\section{Experiments}
In this section, we present experimental results on vectorization and SVG animation. We tested portraits of various styles, including watercolor, painting, manga, and cartoon avatars. Our system supports images of any resolution.

\subsection{Implementation Details}
In order to convert SVG images to raster images, we use a differentiable renderer called DiffVG, and then optimize the path parameters by employing the Adam optimizer. The learning rates for point and color are set to 1 and 0.01, respectively. By default, each path consists of eight segments of third-order Bézier curves. Five levels are exploited to construct the stacke of smoothed images. We adopted 500 paths in total across all the smoothing levels.

All parameters are optimized for 200 iterations in each smoothing level, and 400 iterations in the stage fitting the finest image. For the SVG animation, we use audio-predicted landmarks to drive the SVG portrait to talk. All new frame vector images are deformed from the reconstructed SVG, and the number and order of paths will not change. Additionally, we added eye blinks and head tilts to make the animation more vivid.

\subsection{SVG Reconstruction}
In this section, we created a benchmark containing 20 raster portraits with different styles and resolutions. We then evaluated the SVG Reconstruction of VectorTalker quantitatively and qualitatively, comparing the results with DiffVG and LIVE. DiffVG initializes all paths at once, while LIVE gradually adds new paths, similar to our method. For all experiments, we set the same total number of paths, curve segments, and iterations. As shown in~\cref{fig:compare1}, our method achieved better reconstruction results on various styles of portrait images. Additionally, our method was able to capture complex details that the other methods could not. Even for portraits of real people, we were able to reconstruct an expressive SVG. We highlighted the differences in the boxes on the right, and readers can zoom in for a clearer view.In this section, we created a benchmark containing 20 raster portraits with different styles and resolutions. We then evaluated the SVG Reconstruction of VectorTalker quantitatively and qualitatively, comparing the results with DiffVG~\cite{DiffVG} and LIVE~\cite{LIVE}. DiffVG initializes all paths at once, while LIVE gradually adds new paths, similar to our method. For all experiments, we set the same total number of paths, curve segments, and iterations. Both qualitative~\cref{fig:compare1} and quantitative~\cref{tab:rec} demonstrate that our method achieved better reconstruction results on various styles of portrait images by capturing complex details.

\subsection{SVG animation}
In this section, we present the results of SVG animation. We qualitatively compare ours with MakeItTalk for raster images. Then the vivid control of the SVG portrait is displayed.

We present a comparison of our results with MakeItTalk in~\cref{fig:ComMIT}. Our approach, which uses vector graphics, offers inherent advantages, allowing animations to remain clear and sharp. In contrast, MakeItTalk warps all pixels of the entire image through affine transformation, causing blurry effects and distortions. Additionally, we implemented controlling of eye blink and head poses in addition to the lip motions to make the talking SVG portrait more vivid, as shown i~\cref{fig:BlinkTilt}. 

\begin{table}[htbp]
  \centering
  \begin{tabular}{ccccc}
    \toprule
    Paths & MSE$\downarrow$ & LPIPS$\downarrow$ & PSNR$\uparrow$ & SSIM$\uparrow$ \\
    \midrule
    100 & 0.00498 & 0.350 & 22.856 & 0.772 \\
    250 & 0.00234 & 0.291 & 26.561 & 0.841 \\
    500 & 0.00131 & 0.258 & 29.835 & 0.886 \\
    750 & 0.00112 & 0.222 & 30.085 & 0.893 \\
    1000 & 0.00101 & 0.205 & 30.814 & 0.902 \\
    \bottomrule
  \end{tabular}
  \caption{Ablation on the number of paths. Quantitative evaluation using MSE, LPIPS, PNSR and SSIM on benchmark. More paths will improve reconstructed SVG fidelity and details.}
  \label{tab:table2}
\end{table}

\begin{table}[htbp]
  \centering
  \begin{tabular}{ccccc}
    \toprule
    Levels & MSE$\downarrow$ & LPIPS$\downarrow$ & PSNR$\uparrow$ & SSIM$\uparrow$ \\
    \midrule
    1 & 0.00213 & 0.290 & 27.799 & 0.857 \\
    2 & 0.00163 & 0.285 & 28.524 & 0.872 \\
    3 & 0.00146 & 0.269 & 29.267 & 0.881 \\
    4 & 0.00138 & 0.262 & 29.693 & 0.883 \\
    5 & 0.00131 & 0.258 & 29.835 & 0.886 \\
    \bottomrule
  \end{tabular}
  \caption{Ablation on the number of smoothing levels. Progressive vectorization helps improve reconstruction quality. Note using one level is equivalent to directly using DiffVG.}
  \label{tab:table3}
\end{table}

\begin{figure}[t]
\centering
\includegraphics[width=0.5\textwidth]{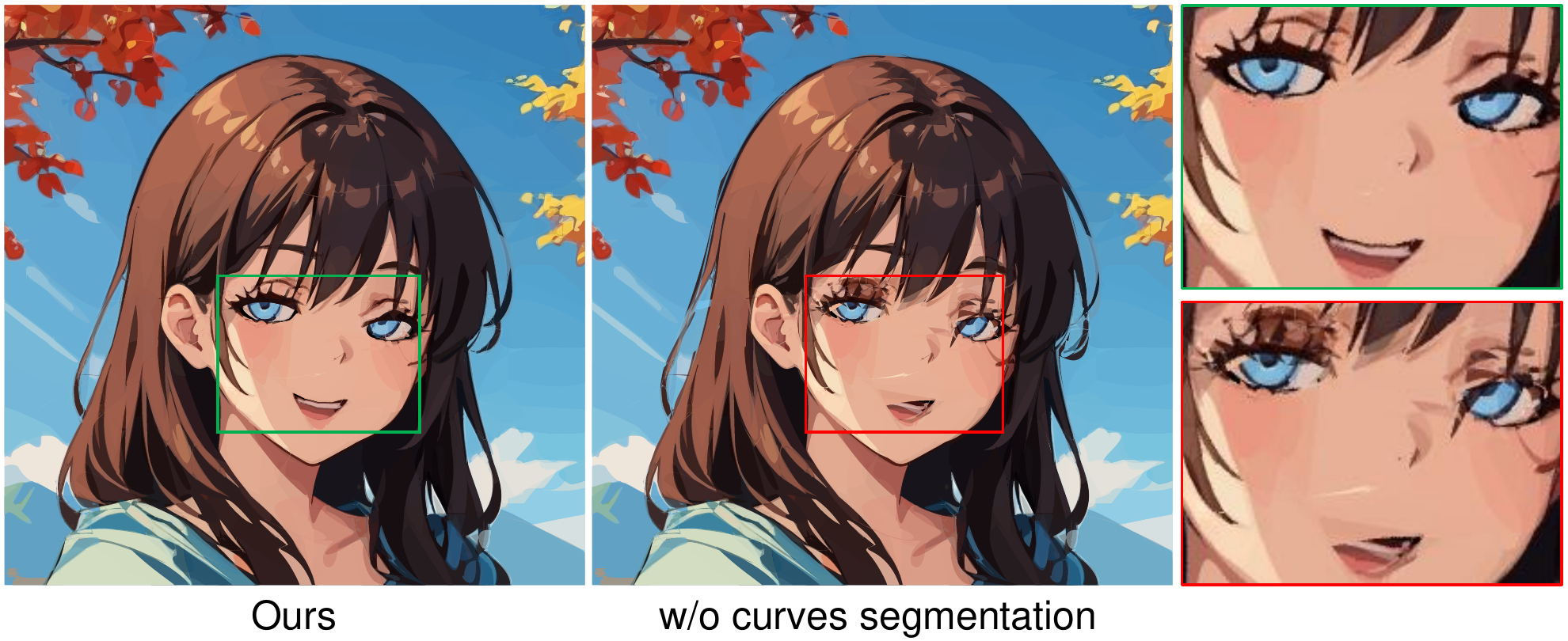}
\caption{Ablation of curves segmentation. The deformed SVG appears messy and distorted without curves segmentation, particularly in the face area where landmarks and triangles are more dense. In contrast, our full method produces accurate results.}
\label{fig:ablation}
\end{figure}

\subsection{Ablation Studies}
We perform several ablation studies on different factors, the number of the paths, the number of the smoothing levels and the effect of the curve segmentation. 

\noindent \textbf{Number of paths.} The total number of paths is a non-differentiable hyperparameter. We ablate the vectorization results using different numbers of paths on our benchmark as shown in~\cref{tab:table2}. We set the paths to 100, 250, 500, 750 and 1000 respectively and keep other parameters the same. 

\noindent \textbf{Number of Levels.} Through progressive vectorization, we apply different numbers of smoothing levels. In~\cref{tab:table3}, we examine the impact of the number of smoothing levels on our benchmarks, using MSE, LPIPS, PNSR, and SSIM with various styles of portraits.

\noindent\textbf{Curve segmentation.}
In order to illustrate whether curves should be split at their intersections with triangulation before warping paths, we compare the animation results of segmented and non-segmented curves. As shown in~\cref{fig:ablation}, the curve segmentation helps to preserve reasonable image structures in animation.

%% file: sec/5_conclusion.tex
\section{Conclusion}

Our research proposes the VectorTalker, a novel approach for generating one-shot audio-driven talking SVG portraits. Our progressive vectorization algorithm allows us to accurately reconstruct the input raster image in vector graphics. We extract facial key points and use an affine-transformation-based warping system to animate the SVG portrait with audio-driven facial key point offset prediction. Our extensive experiments demonstrate that our progressive vectorization significantly outperforms other baseline methods. Additionally, our method effectively accomplishes the task of talking SVG generation. In the future, we plan to utilize more prior knowledge about humans to achieve more vivid control, such as hair and emotion.

%% file: sec/X_suppl.tex
\clearpage
\setcounter{page}{1}
\maketitlesupplementary
\setcounter{figure}{0}

\section*{A. Algorithm}
\cref{alg:alg1} shows the pipeline of VectorTalker, consisting of two stages, progressive vectorization and SVG talking animation. For vectorization, given a reference image, we segment it to get semantic mask and smooth it to obtain different levels of smoothed representations as target images. We progressively add paths and perform the differentiable vectorization to reconstruct three SVG layers and the merged SVG. All results are constrained by current smoothing level target and we optimize the paths by MSE loss. For animation, we extract landmarks to perform triangulation and predict new landmarks from an audio clip. Then we split all the Bezier curves in the SVG at the intersections with the triangulation line segments. Finally, we can warp SVG paths by affine transformation to get animated results.

\begin{algorithm}
\caption{Algorithm of VectorTalker}
\label{alg:alg1}
\begin{algorithmic}[1]
    \STATE \textbf{Input: $I$;} \quad \textcolor{lightgray}{//reference iamge}
    \STATE \textbf{Output: $V$;} \quad \textcolor{lightgray}{//SVG talking animation}
    \STATE \textbf{Procedure:}
    \STATE mask$ = $segment(I)
    \STATE $\{\mathbf{S}^{l}|l=1,...,N\}$ = smooth($I$)
    \STATE para = []; \quad \textcolor{lightgray}{//list of path parameters}
    \STATE errormap = 0;
    \FOR {$i$ in $N$} 
        \STATE newpara = init(errormap,$n$);
        \STATE para = concat(para, newpara);
        \FOR {$j$ in $M$}
            \STATE $\hat{I}$ = render(para);
            \STATE $L$ = loss$(\hat{I}, {S}^{i}, mask)$;
            \STATE para = update($L$, para);
        \ENDFOR
        \STATE errormap = $\left\|{S}^{i}-\hat{I}\right\|_2$
    \ENDFOR

    \STATE $\hat{I}$ = render(para);
    \STATE lmk = detect($\hat{I}$);
    \STATE lmks = pred(lmk, audio);
    \STATE para = segcurves(para, lmk)
    \STATE $V$ = []
    \FOR {$k$ in $K$}
        \STATE para = warp(para, lmks[$k$], lmk)
        \STATE frame = render(para);
        \STATE $V$ = concat($V$, frame)
    \ENDFOR
\end{algorithmic}
\end{algorithm}

\section*{B. Loss Function}
To facilitate SVG animation, we implement a semantic hierarchical design during vectorization to fit three layers respectively. Our loss function, including the MSE loss of each layer and the overall MSE loss of merged SVG, is calculated as follows:

\begin{equation}\begin{gathered}
L_{img}=\lambda_{1}L_{back}+\lambda_{2}L_{fore}+\lambda_{3}L_{local}+\lambda_{4}L_{merged} \\
L_{back}=\left\|inpainting({S}^{l}*mask_{back})-\hat{I}_{back}\right\|_{2}^{2} \\
L_{fore}=\left\|{S}^{l}*mask_{fore}-\hat{I}_{fore}\right\|_{2}^{2} \\
L_{local}=\left\|{S}^{l}*mask_{local}-\hat{I}_{local}\right\|_{2}^{2} \\
L_{merged}=\left\|{S}^{l}-\hat{I}_{merged}\right\|_{2}^{2} \nonumber
\end{gathered}\end{equation}

Where $L_{back}$, $L_{fore}$ and $L_{local}$ are the MSE loss of the background layer, foreground layer, and local layer respectively. ${S}^{l}$ represents the target smoothed image in the l-th level, and $\hat{I}$ represents our rendered result. By default, we set $\lambda_{1}=\lambda_{2}=\lambda_{3}=\lambda_{4}=1$. It is worth mentioning that we keep the background layer fixed in the svg talking animation process. In order to avoid showing a hollow background when head moving, we inpaint the background layer after segmentation.

\begin{figure}[t]
\centering
\includegraphics[width=0.5\textwidth]{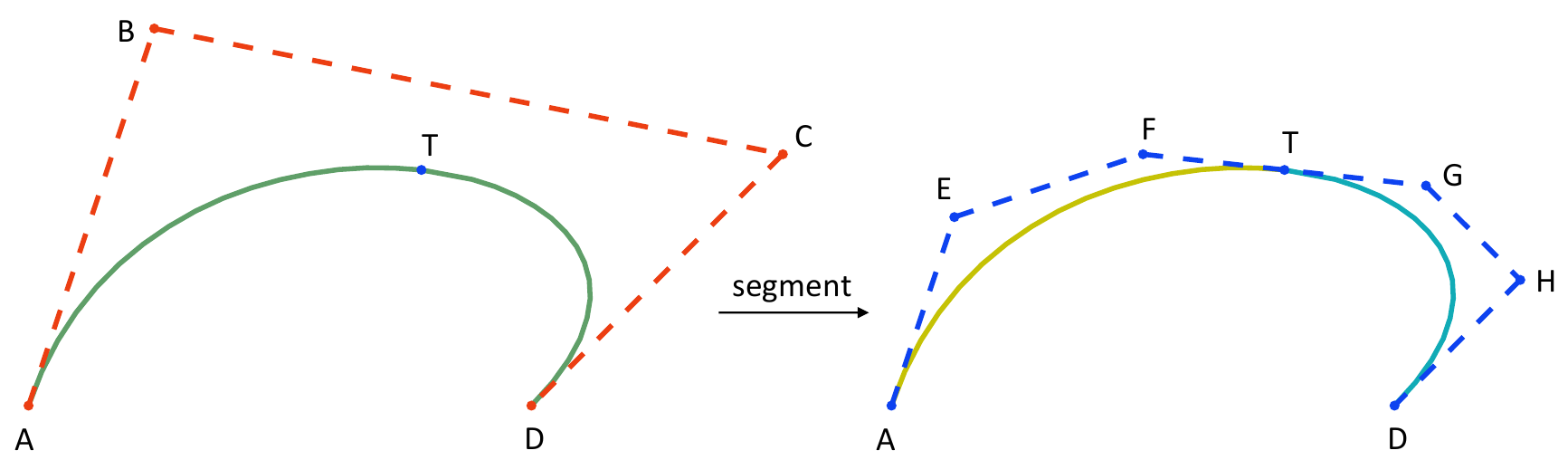}
\caption{Diagram of curve segmentation. The curve is segmented into two at T. The shape will remain unchanged and the number of control points will increase.}
\label{fig:sup}
\end{figure}

\section*{C. Curve Segmentation}
As shown in \cref{fig:sup}, the bezier curve has four control points(A, B, C and D). The point T is on the curve and $T = (1-t)^3A+3(1-t)^2tB+3(1-t)t^2C+t^3D$ where t is the position parameter ranging from 0 to 1. We segment the curve into two parts at T. New control points can be calculated as:

\begin{equation}\begin{gathered}
(A,E,F,T)^{T}=M_0\times (A,B,C,D)^{T} \\
(T,G,H,D)^{T}=M_1\times (A,B,C,D)^{T} \\
M_0=\begin{pmatrix}1&0&0&0\\1-t&t&0&0\\(1-t)^2&2t(1-t)&t^2&0\\(1-t)^3&3t(1-t)^2&3t^2(1-t)&t^3\end{pmatrix} \\
M_1=\begin{pmatrix}(1-t)^3&3t(1-t)^2&3t^2(1-t)&t^3\\0&(1-t)^2&2t(1-t)&t^2\\0&0&1-t&t\\0&0&0&1\end{pmatrix} \nonumber
\end{gathered}\end{equation}

The matrix $M_0$ and $M_1$ can be solved using the method of undetermined coefficients. To segment the curve into multiple parts, we can perform the above operation recursively.